\documentclass{article}

\usepackage[final]{neurips_2021}

\usepackage[utf8]{inputenc}
\usepackage[T1]{fontenc}
\usepackage{hyperref}
\usepackage{url}
\usepackage{booktabs}
\usepackage{amsfonts}
\usepackage{nicefrac}
\usepackage{microtype}
\usepackage{graphicx}
\usepackage{xcolor}
\usepackage{lipsum}
\usepackage{multirow}
\usepackage{adjustbox}
\usepackage{wrapfig}

\title{
  iReason: Multimodal Commonsense Reasoning using Videos and Natural Language with Interpretability\\
}

\author{
  Aman Chadha \\
  Department of Computer Science \\
  Stanford University \\
  \texttt{amanc@stanford.edu} \\
\And
  Vinija Jain \\
  Department of Computer Science \\
  Stanford University \\
  \texttt{vinija@stanford.edu}
}

\begin{document}

\maketitle

\begin{abstract}
   Causality knowledge is vital to building robust AI systems. Deep learning models often perform poorly on tasks that require causal reasoning, which is often derived using some form of commonsense knowledge not immediately available in the input but implicitly inferred by humans. Prior work has unraveled spurious observational biases that models fall prey to in the absence of causality. While language representation models preserve contextual knowledge within learned embeddings, they do not factor in causal relationships during training. By blending causal relationships with the input features to an existing model that performs visual cognition tasks (such as scene understanding, video captioning, video question-answering, etc.), better performance can be achieved owing to the insight causal relationships bring about. Recently, several models have been proposed that have tackled the task of mining causal data from either the visual or textual modality. However, there does not exist widespread prevalent research that mines causal relationships by juxtaposing the visual and language modalities. 
While images offer a rich and easy-to-process resource for us to mine causality knowledge from, videos are denser and consist of naturally time-ordered events. Also, textual information offers details that could be implicit in videos. 
As such, we propose iReason, a framework that infers visual-semantic commonsense knowledge using both videos and natural language captions.
Furthermore, iReason's architecture integrates a causal rationalization module to aid the process of interpretability, error analysis and bias detection. We demonstrate the effectiveness of iReason using a two-pronged comparative analysis with language representation learning models (BERT, GPT-2) as well as current state-of-the-art multimodal causality models. Finally, we present case-studies attesting to the universal applicability of iReason by incorporating the ``causal signal'' in a range of downstream cognition tasks such as dense video captioning, video question-answering and scene understanding and show that iReason outperforms the state-of-the-art.
\end{abstract}



\section{Introduction}
\vspace{-1.5mm}
``On the contrary, Watson, you can see everything. You fail,
however, to reason from what you see.'' \\
\hspace*{\fill} \textit{- Sherlock Holmes, The Adventure of the Blue Carbuncle} \vspace{-1mm}
\noindent\rule{\textwidth}{0.4pt}
\vspace{-3mm}

As humans, a lot is said without explicit connotation \cite{vedantam2015learning, lin2015don}. Humans often possess basic know-how about facts related to the environment we are in and the world at large. For example, if we leave five minutes late, we will be late for the bus; if the sun is out, it’s not likely to rain; and if people are walking on the road, they're using their legs to do so. Humans learn commonsense in an unsupervised fashion by exploring the physical world, and until machines imitate
this learning path by imbibing the contextual property of causal knowledge in their understanding, there will be an inevitable ``gap'' between man and machine.

The aforementioned implicit knowledge fosters \textit{commonsense of causality} \cite{pearl2018book} in everyday life. Causality helps identify the cause-and-effect relationship between events, which enables gaining deeper insights about not only the casual connections between the events themselves but also of the environment in which these events occur. This has the effect of improving the understanding of the happenings in a real-life events depicted through a video or a natural language snippet, not just for humans but also for deep-learning models \cite{chadha2020iperceive, wang2020causal}.

Prior work \cite{hendricks2018women, manjunatha2019explicit} has unraveled spurious observational biases that models fall prey to in the absence of causality. Causal relationships remedy this by helping point out contextual attributes - such as if there's \textit{barking} noise, a \textit{dog} should be present - that are usually implied for humans. The problem of causality-in-AI \cite{chadha2020iperceive} thus has broad applicability to a wide gamut of vision and text-based tasks. Specifically, causality-in-AI can help improve the robustness of downstream tasks that suffer from limited performance owing to the lack of understanding causal relationships such as dense video captioning \cite{krishna2017dense, zhou2018end, wang2018bidirectional}, 
video question-answering \cite{zeng2017leveraging, hu2017learning, antol2015vqa, singh2019towards}, and a plethora of other NLP tasks \cite{oh2013question, hashimoto2014toward, ning2019joint, zhang2020learning, rajani2019explain, chadha2020iperceive}, etc. 
This makes it valuable to impart the notion of causality to machines.

While most other work in the domain of causality-in-AI requires expensive hand annotation \cite{fang2020video2commonsense, rajani2019explain, wang2020causal} to acquire commonsense knowledge owing to their exclusive use of the text modality, relatively little work exists in literature that utilizes visual modalities. On the flipside, while there is work in the field \cite{chadha2020iperceive} that generates commonsense using self-supervised methods (thus obliterating the need for expensive hand annotation), but it is limited to the visual modality (and thus doesn't imbibe learnings from natural language snippets -- say captions -- using NLP). Zhang et al. \cite{zhang2020learning} propose a unique direction in causality-in-AI by proposing a vision-contextual causal (VCC) model that utilizes both the visual and language modality to infer commonsense knowledge. However, under the visual modality, \cite{zhang2020learning} limits commonsense knowledge generation from images and cannot natively accept videos as input, which are a much more prevalent source of commonsense knowledge compared to images and text. Instead, the model utilizes a pair of randomly-selected images to be able to infer commonsense knowledge which limits its effectiveness.

Furthermore, the ability to \textit{rationalize} causal relationships is instrumental to not only ease the process of error analysis but also instill confidence in the model's predictions. VCC \cite{zhang2020learning} doesn't offer rationalization for its causality inference, thereby hindering the model's interpretability. This also makes it difficult to understand the source of error/bias when analyzing results since rationalization can offer a peak into the model's modus operandi, i.e., act as a `debug signal' to develop an understanding of the model's (mis)learnings.

To push the envelope of causality-in-AI, we propose iReason, a framework that generates commonsense knowledge by inferring the causal relationships using two of the most knowledge-rich modalities -- videos and text. This enables the model
to seek intrinsic causal relationships between objects within
events in a video sequence and supplement the knowledge thus gained using natural language snippets, i.e., captions of the aforementioned events. To demonstrate that iReason furthers the state-of-the-art, we offer hands-on evaluation by comparing our results to textual representation learning models (BERT, GPT-2) in addition to the current state-of-the-art causality models. Furthermore, we present case-studies by incorporating the ``causal signal'' in downstream cognition tasks such as dense video captioning and video question-answering and show that imbibing causality knowledge using iReason into the aforementioned tasks helps them outperform the current state-of-the-art.

In summary, our key contributions are centered around the following.
\begin{enumerate}
    \item \textbf{Commonsense reasoning using videos and natural language:} iReason infers causal knowledge grounded in videos and natural language. We envision this as a step towards human-level causal learning. As such, iReason as a dual-grounded causality learning approach offers the following advantages:
    \begin{enumerate}
        \item \textbf{Causality using the visual and text modality:} Videos prevalently contain commonsense knowledge that cannot be easily inferred using just text because such information is not usually explicitly specified in textual form \cite{pearl2018book}. For e.g., consider a video of a girl throwing a frisbee in the air (event $X$) and a dog jumping to catch it (event $Y$). In this case, there exists a causal relationship between the two events (event $X$ $\rightarrow$ event $Y$). While a textual caption of the entire sequence would be helpful in understanding the events, it would typically fail to explicitly specify this relationship. However, the fact that the girl threw a frisbee (event $X$) \textit{led} to the dog jumping (event $Y$) would be apparent from the video.
        As such, both modalities hold their unique importance in the task of learning causal relationships. iReason thus seeks to blend both the visual and text modality to mine commonsense. Figure \ref{causal} traces this example through iReason.
        \item \textbf{Exploit the time-ordered nature of videos:} There exists a strong correlation between temporal and causal relations (say $A$ is the cause and $B$ is the effect, then $A$ has to precede $B$ in time). Since events in most video sequences are naturally time-ordered, they are an apt 
        resource for us to mine cause-and-effect relationships from.
        \item \textbf{Use objects in videos to develop an idea of contextual causality:} Objects in videos can be used as environmental context to understand causal relations in the scene.
    \end{enumerate}
    \item \textbf{Offer interpretability and error detection:} iReason can rationalize causal relationships and thus help us understand its learnings using natural text. This would help perform error analysis and more importantly, also spot biases in the dataset.
    \item \textbf{Universal applicability to cognition tasks:} iReason's commonsense features can be incorporated in downstream tasks that require cognition by supplementing them with the input features (cf. Section \label{downstream}). It is thus noteworthy that they have a certain universality and are not limited to the realizations of DVC and VideoQA discussed in this work. As such, they can be easily adapted for other video-based vision tasks such as scene understanding \cite{hu2020probabilistic}, panoptic segmentation \cite{kim2020video}, etc.
\end{enumerate}

iReason thus infers visual-semantic causal knowledge by blending videos and natural language to perform multimodal commonsense reasoning. This is accomplished by localizing events in videos, drawing on canonical frames that represent these events, and learning contextual causal relationships using both videos and text captions. Furthermore, iReason's architecture integrates a causal rationalization module to aid the process of interpretability, error analysis and bias detection. 

\section{Related Work}

Inferring causal relationships to bolster machine intelligence has been an area that has been under the spotlight in recent times owing to it being a significant step towards artificial general intelligence (AGI) \cite{pearl2018book, goertzel2014artificial, scholkopf2019causality}. Commonsense knowledge can be derived using either the language and/or visual (images or video) modality and  literature in the field can thus be reviewed in a similar fashion.

\subsection{Causality in Natural Language}
 
Several approaches \cite{storks2019commonsense, aditya2015images, sap2020commonsense, fang2020video2commonsense, rajani2019explain, wang2020causal} have been proposed that extract causal knowledge using natural language. These approaches either mine textual snippets such as captions, text blurbs or large-scale knowledge bases such as Wikipedia. However, causality grounded in natural language has the obvious disadvantage of being limited by the reporting bias \cite{vedantam2015learning, lin2015don} (for e.g., washing a car \textit{leads to} the car being clean is something that is not explicitly mentioned in text, but can easily be visually inferred) and thus suffer from sub-standard performance.

\subsection{Causality in Vision}
There has been a recent surge of interest in coupling the complementary strengths of computer vision and causal reasoning \cite{pearl2016causal, pearl2014interpretation}. The union of these fields has been explored in several contexts, including image classification \cite{lopez2017discovering, chalupka2014visual}, reinforcement learning \cite{nair2019causal}, adversarial learning \cite{kocaoglu2017causalgan}, visual dialog \cite{qi2019two}, image captioning \cite{zhou2020more} and scene/knowledge graph generation \cite{tang2020unbiased, pan2020spatio}. While these methods offer limited task-specific causal inference, current research that tackles the task of building a generic commonsense knowledge base from visual input mainly falls into two categories: (i) learning from images \cite{yatskar2016stating, vedantam2015learning, zhu2014reasoning, wang2020visual} and (ii) learning actions from videos \cite{goyal2017something}. While the former limits learning to human-annotated knowledge which restricts its effectiveness and outreach, the latter is essentially learning from correlation.

\subsection{Multimodal Causality}

With the recent success of pre-trained language models \cite{devlin2018bert, dai2019transformer, peters2018deep} in NLP, several approaches \cite{lu2019vilbert, sun2019videobert, tan2019lxmert, chen2020uniter} have emerged that utilize weakly-supervised learning models using large, unlabelled, multimodal (images and natural language) data to encode visual-semantic knowledge. However, the inordinate memory cost for 
task-specific finetuning is a significant 
barrier-to-entry 
for 
such systems.

\section{Task Definition}

The goal of this work is to mine contextual causality knowledge from videos and natural language. We formally define the task as follows:

\begin{enumerate}
    \item The input to the model is a video, fed to the canonical frame identification module, which outputs an image pair $P \in \mathcal{P}$, where $\mathcal{P}$ is the set of image pairs from each event $e \in E$, where $E$ is the set of all events in the video. $P$ thus consists of two frames $I_{1}$ and $I_{2},$ sampled from the video $V$, in temporal order (i.e., $I_{1}$ appears before $I_{2}$, and thus $I_{1}$ and $I_{2}$ are the cause and effect frames respectively).
    \item For each $P$, our goal is to identify all possible causal relations between $I_1$ and $I_2$. Normally, this task contains two sub-tasks: (i) identifying events in frames and, (ii) identifying causal relations between the said events. 
    \item The canonical frame identification module (cf. Section \ref{arch}) enables the first sub-task. For the second sub-task, we assume that the set of events contained in $I_{1}$ is denoted as $\mathcal{E}_{1}$ and the set of events contained in all frames sampled from $V_{1}$ is denoted as $\mathcal{E}_{v}$. For each event $e_{1} \in \mathcal{E}_{1},$ our goal is finding all events $e_{2} \in \mathcal{E}_{v}$ such that $e_{1}$ causes $e_{2}$ ($e_{1}$ $\rightarrow$ $e_{2}$).
    \item The output of the model is a causality score prediction $C \in [0, 1]$ interpreted as a probability measure. By setting a compliance threshold for $c$ (say, 0.5), positive ($e_{1}$ $\rightarrow$ $e_{2}$) and negative ($e_{1}$ $\not\rightarrow$ $e_{2}$) causal relationships can be inferred.
    \item Finally, we rationalize our causality output using the causality rationalization module which accepts the output $c$ from the prior step along with $e_{1}$, $e_{2}$ and the outputs a string explaining our rationale behind the prediction.
\end{enumerate}

\begin{wrapfigure}{R}{0.45\linewidth}
    \centering
    \includegraphics[width=0.98\linewidth]{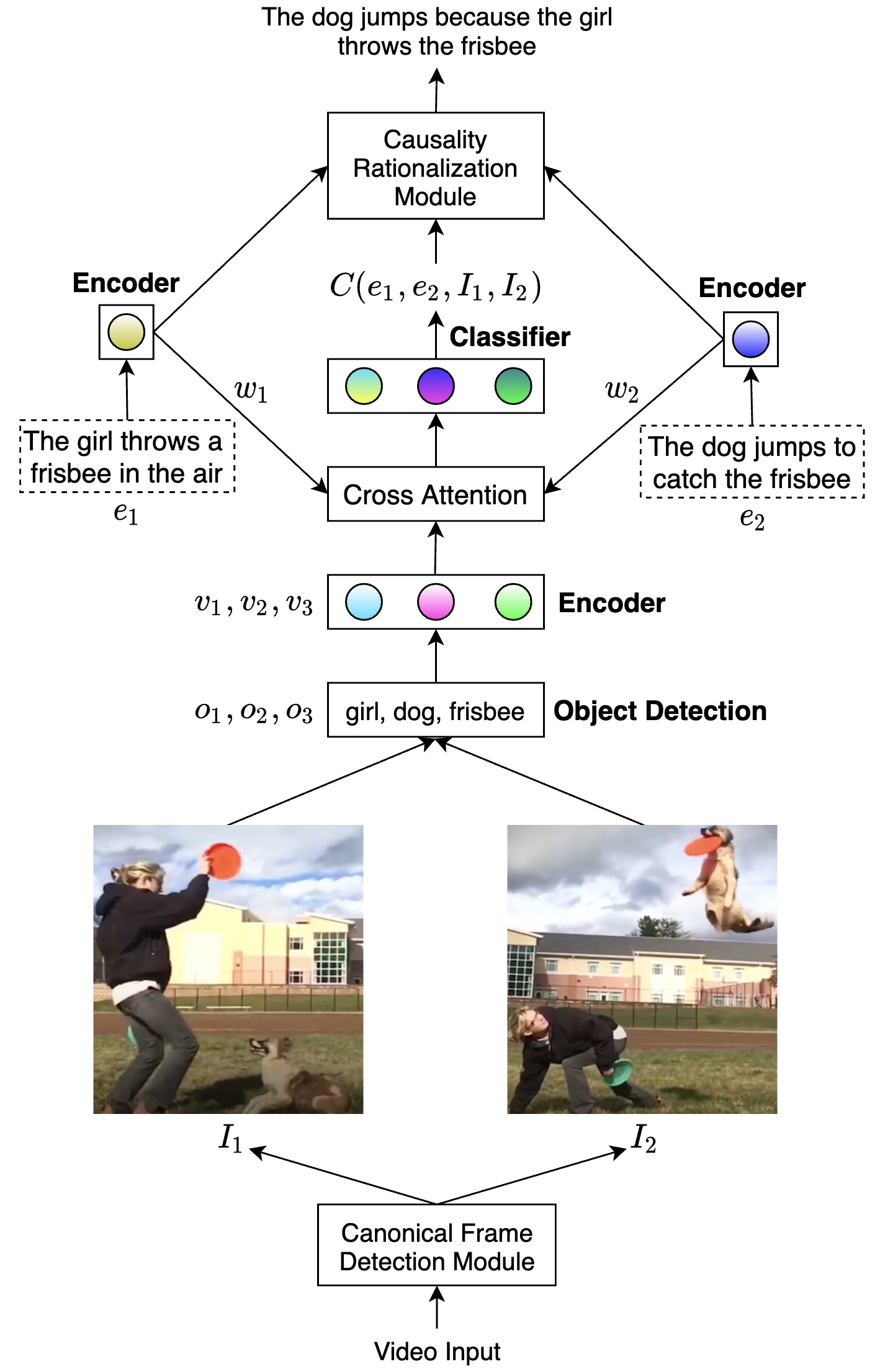} 
    \vspace {-2mm}
    \caption{Architectural overview of iReason.}
    \label{causal}
\end{wrapfigure}

\section{Network Architecture} \label{arch}

iReason is an end-to-end trainable model that discerns causal relationships between temporal events in videos. To this end, we propose the following architecture as in figure \ref{causal}: (i) the canonical frame detection module, which accepts video input, localizes events in the said videos and identifies representative frames corresponding to a pair of events; (ii) the textual encoder module, which encodes labels of detected objects from a pair of contextual frames into vectors; (iii) the cross-attention module, where the resultant features are attention-intertwined to find context and event representation; (iv) a binary classifier, which outputs the causal probability prediction which is ultimately fed (along with the encoded event captions) to the causality rationalization module to foster model interpretability.

\subsection{Canonical Frame Detection Module}
While VCC \cite{zhang2020learning} utilizes a pair of event frames as input to the model, they are sampled at uniform intervals and the events corresponding to the images are thus essentially randomly picked (since there is little correlation between the timing of event occurrences in different videos, say a dog jumping \textit{in response to} a frisbee thrown in the air vs. the opening of a door \textit{as a result of} the doorbell ringing). Statically selected event frames are thus error-prone, do not accommodate overlapping events and are limited in magnitude (2-3 events per video on an average). To remedy this, we utilize the event localization module in \cite{chadha2020iperceive} to derive a set of events from the input video, which are much more exhaustive (7-10 events per video on an average) and contain both overlapping and non-overlapping events. Furthermore, we propose a canonical frame detection algorithm which enables the model to pick a representative frame given an event (vs. statically chosen as in \cite{zhang2020learning}). 

The canonical frame detection algorithm performs activity detection on frames within a particular localized event to infer visual context. We do so using a pre-trained model from \cite{liu2018t}. Next, we perform a unigram-based BLEU score \cite{papineni2002bleu} match to identify the first frame with the best match of constituent objects compared to those in the event caption. Formally,

\begin{equation}
BLEU=BP \times \exp \left(\lambda_{n} \log p_{n}\right)
\end{equation}

where, BP is the brevity penalty which is set to 1.

This enables iReason to natively accept videos, learn deeper causal relationships and facilitates end-to-end training with video input. Our canonical frame detection module thus has the distinct novel advantage of being able to establish causal relationships between a broad range of events -- overlapping/non-overlapping, short/long, few/many, etc.

\subsection{Textual Encoder Module}\label{objdet}

Similar to \cite{zhang2020learning}, we use BERT \cite{devlin2018bert} to encode textual representations of events $e_1$ and $e_2$, denoted by $w_1$ and $w_2$.
Following scene-graph approaches \cite{xu2017scene, abhishek2017devi}, we leverage a pre-trained Faster R-CNN \cite{ren2015faster} model trained on MS-COCO \cite{lin2014microsoft}, to perform object detection on the canonical frames and thus establish visual context. Next, we choose the top $m$ object predictions sorted by their confidence score, where $m$ is a hyperparameter (cf. Section \ref{hyperparam}). Finally, we encode the vector representations of the selected objects $o_1$, $o_2$, $\ldots$, $o_n$ using BERT and denote them as $v_1$, $v_2$, $\ldots$, $v_n$.

\subsection{Cross-Attention Module}

The cross-attention module seeks to select (i) objects associated with events thus establishing context, and (ii) events associated with the aforementioned context.

\textbf{Context Representation.} For each event $e,$ whose tokens' vector representations are ${w}_{1}, {w}_{2}, \ldots, {w}_{n},$ we first take the average of all tokens and denote the resultant average vector as $\widetilde{{w}}$. Next, with the set of all objects denoted as $\mathcal{O}$, we compute the context representation as:

\begin{equation}
{o} = \sum_{{o}^{\prime} \in \mathcal{O}} a_{\widetilde{{w}}, {o}^{\prime}} \cdot {o ^ { \prime }}
\end{equation}
$$a_{\widetilde{{w}}, {o}^{\prime}}=NN_{a}\left(\left[\tilde{{w}}, {o}^{\prime}\right]\right)$$

where $a_{\widetilde{{w}}, {o}^{\prime}}$ is the attention weight of $\widetilde{{w}}$ on object $o^{\prime}$ computed by a two-layer feed forward neural network $NN_{a}$ and $[,]$ indicates concatenation.

\textbf{Event Representation.} 
Assuming that the set of events $e$ is denoted by $\mathcal{W}$, we can get the event representation using a similar attention structure:

\begin{equation}
{e} =\sum_{{w}^{\prime} \in \mathcal{W}} b_{{o}, {w}^{\prime}} \cdot {w}^{\prime}
\end{equation}
$$
b_{{o}, {w}^{\prime}} =NN_{b}\left(\left[{o}, {w}^{\prime}\right]\right)
$$

where $b_{{o}, {w}^{\prime}}$ is the attention weight computed by another feed forward neural network $NN_{b}$.

\subsection{Causality Score Prediction}

Assuming that the context representations for $e_{1}$ and $e_{2}$ are denoted as ${o}_{e_{1}}$ and ${o}_{e_{2}}$ respectively, we can predict the final causality score using a binary classifier $NN_c$ as:


\begin{equation}
C\left(e_{1}, e_{2}, I_{1}, I_{2}\right)=N N_{c}\left(\left[{w}_{1}, {w}_{2}, {o}_{e_{1}}, {o}_{e_{2}}\right]\right)
\end{equation}

\subsection{Causality Rationalization Module}
We enhance the interpretability and robustness of iReason by adapting the commonsense auto-generated explanations (CAGE) framework proposed by Rajani et al. \cite{rajani2019explain} which generates explanations for commonsense reasoning using natural language (cf. Section \ref{dataset}) for causal events $e_1$ and $e_2$. Given a question $q$, four answer choices $c_0$, $c_1$, $c_2$, and a labeled answer $a$, CAGE generates explanations $e_i$ according to a conditional language modeling objective as follows:

\begin{equation}
- \sum_{i} \log P\left(e_{i} \mid e_{i-k}, \ldots, e_{i-1}, C_{txt} ; \theta\right)
\end{equation}

where, $C_{txt}$ is the input context defined as ``$q, c_0, c_1$ or $c_2$? commonsense says <explanation>'' and $k$ is its size; $\theta$ is the set of parameters conditioned on $C_{txt}$ and previous explanation tokens; 

We adapt CAGE for iReason by forming a question using $e_2$ 
by prepending it with ``Why does $\ldots$''. Next, we perform activity detection using a pre-trained model from \cite{liu2018t} on 
$I_1$ and fetch the top three ranked activities.
Table \ref{cage} offers insights into CAGE's inputs/outputs using the example in figure \ref{causal}.


\begin{table}[h]
    \caption{Adapting CAGE for iReason.}
    \vspace {-2.5mm}
    \centering    
    \label{cage}
    \setlength{\tabcolsep}{4.5pt}
\scriptsize
    \begin{tabular}{ccc}
        \toprule
\multirow{2}{*}{Input} & Question & Why does the dog jump to catch the frisbee? \\
& Choices & girl throws frisbee, dog sits on grass, girl stands on grass \\
\hline\noalign{\vskip 0.6mm}
Output & CAGE & The dog jumps because the girl throws the frisbee. \\
\bottomrule
    \end{tabular}
\end{table}

\subsection{Loss Function}



For each positive example in the Vis-Causal dataset \cite{zhang2020learning}, we randomly select one negative example and use cross-entropy as the loss function. Formally,

\begin{equation}
J = CrossEntropy\left(I_i^{\prime}, I_j\right)
\end{equation}

where, $I^{\prime}$ is the $i^{th}$ positive sample and $I_i$ is the $j^{th}$ randomly selected negative sample.








\section{Experiments}

\subsection{Data}\label{dataset}

For learning causal relationships, we propose iReasonData, a dataset that amalgamates videos from the ActivityNet dataset \cite{caba2015activitynet} (which contains short videos from YouTube) and causal annotations from the Vis-Causal dataset \cite{zhang2020learning} (which contains causal relationships between events in frames sampled from the aforementioned videos). This enables us to run the canonical frame detection module directly on the ActivityNet videos which serve as an input to iReason, thereby enabling iReason to select representative frames from the selected events. 
The output is a binary label indicating the causal relationship between the events pictured in the sampled canonical frames. We chose the same 1,000 videos obtained from the ActivityNet dataset \cite{caba2015activitynet} that Vis-Causal used and the corresponding causal annotations from Vis-Causal for the videos. Our training/validation/test split was 80\%/10\%/10\%. Table \ref{ireasondata} presents details of the iReasonData. 

\begin{table}[h]
    \caption{Specifics of the iReasonData dataset.}
    \vspace {-2.5mm}
    \centering    
    \label{ireasondata}
    \setlength{\tabcolsep}{4.5pt}
\scriptsize
    \begin{tabular}{ccc}
        \toprule
\multirow{5}{*}{Input} & \# of videos (from the ActivityNet dataset) & 1,000\\
& \# of frames sampled from each video & 4 \\
& Total \# of frame pairs & 1,000 (videos) $\times$ 4 (frames/video) = 4,000\\
& \# of event annotations for each video & 3\\
& Total \# of event annotations & 4,000 (videos) $\times$ 3 (annotations/video) = 12,000\\
\hline\noalign{\vskip 0.6mm}
Output & \# of causal relationships & 12,000 (annotations) $\times$ 1 (causal relationship) = 12,000\\
        \bottomrule
    \end{tabular}
\end{table}


For learning causality rationalization, we used the pre-trained model in \cite{rajani2019explain} trained on the CoS-E dataset, which consists of 9741 question-choices-answer samples (with a training/validation/test split of 80\%/10\%/10\%). Table \ref{cosedata} explores the structure of a sample within CoS-E. 

\begin{table}[h]
    \caption{Structure of a sample within the CoS-E dataset.}
    \vspace {-2.5mm}
    \centering    
    \label{cosedata}
    \setlength{\tabcolsep}{4.5pt}
\scriptsize
    \begin{tabular}{cc}
        \toprule
\multirow{1}{*}{Input} & Question; three answer choices; ground-truth answer\\
\hline\noalign{\vskip 0.6mm}
\multirow{2}{*}{Output} & Highlighted relevant words in the question that justify the ground-truth answer\\
& Brief explanation based on the highlighted justification that serves as the commonsense reasoning \\
        \bottomrule
    \end{tabular}
\end{table}

\subsection{Evaluation method}

Since each event in $I_1$ could cause multiple events in $I_2$, we evaluate different causality extraction models (cf. Section \ref{secresults}) with a ranking-based evaluation metric \cite{xu2017scene}. Given an event $e$ in $I_1$, models are required to rank all candidate events based on how likely they think these events are caused by $e$. We then evaluate different models using 
Recall@N (R@N),
where $N$ denotes whether the correct causal event is covered by the top one (R@1), five (R@5), or ten (R@10) ranked events.

\subsection{Experimental Details} \label{hyperparam}

Table \ref{config} presents details of the training knobs and hyperparameters \cite{zhang2020learning}.

\begin{table}[h]
    \caption{iReason's training config.}
    \vspace {-2.5mm}
    \centering    
    \label{config}
    \setlength{\tabcolsep}{4.5pt}
\scriptsize
    \begin{tabular}{cc}
        \toprule
Optimizer & Stochastic gradient descent\\
Parameter initialization & Random \\
Learning rate & $10^{-4}$\\
Number of epochs & 10 \\
Early stopping & Yes \\
Number of detected objects considered & 10 \\
Hidden layer size in the feed-forward neural net & 200 \\
Total trainable parameters & 112.1 million (including 109.48 million from BERT-base) \\
        \bottomrule
    \end{tabular}
\end{table}

\subsection{Results} \label{secresults}

Table \ref{results} compares iReason for the various context categories with VCC \cite{zhang2020learning}, while table \ref{langresults} compares iReason with language representation learning models (BERT \cite{devlin2018bert}, GPT-2 \cite{radford2019language}). Table \ref{viz} offers a visual walkthrough of iReason with positive and negative examples.

\subsection{Ablation Experiments}

To prove that the multimodal context is crucial for learning causality, we carried out the following ablation experiments.
Table \ref{results} offers an in-depth view into these experiments.
\begin{enumerate}
    \item \textbf{No lingual context:} Predict causal relationships without using the language modality, i.e., limit the model to only use videos.
    \item \textbf{No visual context:} Predict causal relationships without using videos, i.e., limit the model to only use natural language captions.
\end{enumerate}

\begin{table}[!hb]
    \caption{Performance analysis of iReason and ablated versions compared to VCC \cite{zhang2020learning}. 
    Bold numbers indicate best performance.}
    \vspace {-1.5mm}
    \centering    
    \label{results}
    \setlength{\tabcolsep}{4.5pt}
\scriptsize
\begin{tabular}{lcccccccc}
        \toprule 
        Model & Metric & Sports & Socializing & Household & Personal Care & Eating & Overall \\
\hline
\noalign{\vskip 0.6mm}
& R@1 & 0.67 & 3.64 & 1.69 & 0.00 & 9.09 & 2.13 \\ 
Random guess & R@5 & 14.19 & 16.36 & 15.25 & 11.11 & 27.27 & 15.25 \\
& R@10 & 28.38 & 38.18 & 27.12 & 33.33 & 27.27 & 30.14 \\
\hline
\noalign{\vskip 0.6mm}
\multirow{1.8}{*}{No lingual context} & R@1 & 7.47 & 7.17 & 6.28 & 10.88 & 24.55 & 7.17 \\
\multirow{-0.5}{*}{(use only videos)} & R@5 & 35.15 & 34.12 & 27.12 & 31.22 & 39.11 & 30.64 \\
& R@10 & 60.27 & 54.18 & 58.15 & 52.11 & 66.22 & 58.23 \\
\hline
\noalign{\vskip 0.6mm}
\multirow{1.8}{*}{No visual context} & R@1 & 7.88 & 7.66 & 6.45 &11.01 & 24.57 & 8.13 \\
\multirow{-0.5}{*}{(use only natural language captions)} & R@5 & 36.11 & 34.36 & 28.77 & 32.17 & 42.77 & 31.22 \\
& R@10 & 61.11 & 55.15 & 58.22 & 52.25 & 67.73 & 60.14 \\
\hline
\noalign{\vskip 0.6mm}
& R@1 & 8.78 & 7.27 & 6.78 & 11.11 & 27.27 & 8.87 \\
VCC \cite{zhang2020learning} & R@5 & 37.16 & 36.36 & 28.81 & 33.33 & 45.45 & 34.75 \\
& R@10 & 64.86 & 58.18 & 62.71 & 55.56 & 72.73 & 63.12 \\
\hline\noalign{\vskip 0.6mm}
& R@1 & $\textbf{9.27}$ & $\textbf{8.09}$ & $\textbf{7.91}$ & $\textbf{12.72}$ & $\textbf{28.89}$ & $\textbf{9.21}$ \\
iReason & R@5 & $\textbf{38.71}$ & 36.36 & $\textbf{29.92}$ & $\textbf{34.73}$ & $\textbf{45.75}$ & $\textbf{35.87}$ \\
& R@10 & $\textbf{65.12}$ & $\textbf{58.52}$ & 62.71 & $\textbf{55.86}$ & 72.73 & $\textbf{63.51}$ \\
        \bottomrule
    \end{tabular}
\end{table}

\renewcommand{\arraystretch}{1}
\begin{table}[!hb]
    \caption{Performance comparison with BERT and GPT-2. 
    Bold numbers indicate best performance.}
    \vspace {-1.5mm}
    \centering    
    \label{langresults}
    \setlength{\tabcolsep}{4.5pt}
\scriptsize
\begin{tabular}{lccc}
        \toprule
 & R@1 & R@5 & R@10 \\
\midrule 
Random Guess & 2.13 & 15.25 & 30.14 \\
BERT & 2.13 & 22.34 & 39.00 \\
GPT-2 & 3.55 & 17.73 & 34.40 \\
\midrule
VCC (BERT) & {8.87} & {34.75} & {63.12} \\
VCC (GPT-2) & 7.80 & 31.56 & 56.03 \\
\midrule 
iReason (BERT) & $\textbf{9.21}$ & $\textbf{35.87}$ & $\textbf{63.51}$ \\
iReason (GPT-2) & 8.90 & 35.21 & 63.23 \\
\bottomrule
    \end{tabular}
\end{table}

\begin{table}[!hb]
    \caption{Visually inspecting positive and negative examples from iReason.}
    \vspace {-3.5mm}
    \centering    
    \label{viz}
    \setlength{\tabcolsep}{4.5pt}
\scriptsize
        \begin{center}
\begin{tabular}{cccc}
        \toprule 
        Canonical frame detection output ($I_1$, $I_2$) & Event captions & Causal prediction & Causality rationalization output \\
\hline
\noalign{\vskip 1.5mm}
\multirow{2}{*}{\adjustbox{valign=m}{\resizebox{30pt}{!}{\includegraphics{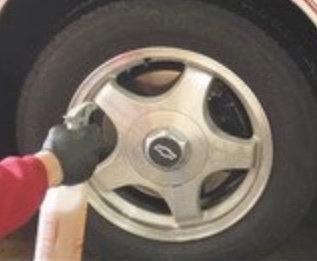}}}} \multirow{2}{*}{\adjustbox{valign=m}{\resizebox{30pt}{!}{\includegraphics{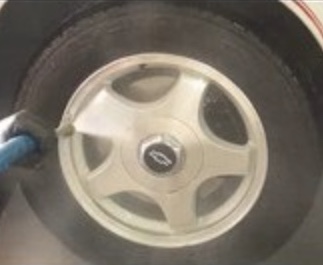}}}} & \multirow{1.1}{*}{$e_1$: Tire shine is being applied.} & \multirow{2}{*}{Yes} & \multirow{1.1}{*}{The rims are shiny because} \\
& $e_2$: The rims are shiny. & & tire shine was applied. \\
\noalign{\vskip 4mm}
\multirow{2}{*}{\adjustbox{valign=m}{\resizebox{30pt}{!}{\includegraphics{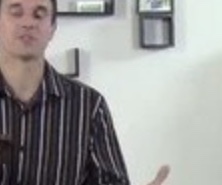}}}} \multirow{2}{*}{\adjustbox{valign=m}{\resizebox{30pt}{!}{\includegraphics{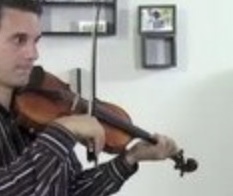}}}} & \multirow{1.1}{*}{$e_1$: A man is talking.} & \multirow{2}{*}{No} & \multirow{1.2}{*}{No causal rationalization since} \\
& $e_2$: A man plays the violin. & & events are not causal.\\
\noalign{\vskip 3.5mm}
\bottomrule
    \end{tabular}
            \end{center}
\end{table}

\section{Analysis of Results}

\begin{enumerate}
    \item From table \ref{results}, we observe that iReason outperforms VCC \cite{zhang2020learning} on R@1 across-the-board, and on R@5 and R@10 for most categories (while still maintaining the performance levels of VCC on R@5 and R@10 for the other categories). The fact that iReason outperforms VCC on R@1 implies that iReason is much more effective than VCC to identify the correct causal event using only the first ranked event. Furthermore, in some context categories under R@5 and R@10, VCC offers the same performance levels as iReason (which implies requiring atleast five and ten of the top ranked events respectively to select the correct causal event) since the additional events increase the probability of VCC of identifying the right answer. 
    \item From table \ref{results}, iReason outperforms the ``no visual/lingual context'' ablated models for all settings, which proves the importance of visual-semantic context. Also, since using only lingual context outperforms using only visual context, lingual context is more useful than visual context to learn commonsense knowledge. However, they offer complementary knowledge in most cases leading to iReason performing better than either ablated model.
    \item From table \ref{langresults}, we see that all models significantly outperform the ``random guess'' baseline in almost all settings, which shows that the models have learned to extract meaningful semantic and/or causal knowledge from their respective input modalities. 
        \begin{enumerate}
        \item Compared with the ``random guess'' baseline, BERT and GPT-2 achieve slightly better performance. This implies that even though these pre-trained language representation models encode semantic information about events, they can only identify the probabilistic relevance between the events, rather than their causal relationship.
        \item iReason outperforms VCC \cite{zhang2020learning} which indicates that canonical frame detection helps mine additional causal knowledge compared to VCC. 
        \end{enumerate}
\end{enumerate}

\section{Conclusion}
We proposed iReason, a multimodal causal knowledge generation framework. iReason outperforms not only language representation models but also state-of-the-art causal models owing to its canonical frame detection module (which enables video as native inputs, rather than randomly-selected frames) and causal rationalization module (which offers interpretability). With ablation experiments, we assessed the role played by multimodal context.
To improve iReason, a couple of ideas can be explored. iReason's performance is a function of the object detection module and the language representation model. Using a better object detector and language representation model could improve context availability and event coding respectively for iReason and thus offer enhanced performance. 
Furthermore, to even remotely match the magnitude of a human-level commonsense knowledge base requires a dataset 
that covers a multitude of scenarios. While iReasonData is effective, more data is needed to cover additional causal interactions that humans easily infer (sub)consciously.
\bibliographystyle{unsrt}
\bibliography{references}



\end{document}